\documentclass[alpha-refs]{wiley-article}

\usepackage{siunitx}
\usepackage{hyperref}
\usepackage{blindtext}
\usepackage{graphicx}
\usepackage{fix-cm}
\usepackage{lmodern}
\usepackage[T1]{fontenc}
\DeclareFontShape{T1}{lato-TLF}{b}{sc}{<->ssub*lato-TLF/b/n}{}

\papertype{Original Article - Draft}

\paperfield{}

\title{Artificial Intelligence Mangrove Monitoring System Based on Deep Learning and Sentinel-2 Satellite Data in the UAE (2017-2024)}

\author[1\authfn{1}]{Linlin Tan}
\author[2\authfn{1}]{Haishan Wu}

\contrib[\authfn{1}]{Corresponding authors.}

\affil[1]{Ubiquitous Sensing and Intelligent Detection Team, Guangdong Laboratory of Artificial Intelligence and Digital Economy (SZ), Shenzhen, 518000, China}
\affil[2]{Zand Bank, PJSC, Dubai, United Arab Emirates}

\corraddress{Ubiquitous Sensing and Intelligent Detection Team, Guangdong Laboratory of Artificial Intelligence and Digital Economy (SZ), Shenzhen, 518000, China}
\corremail{tanlinlin@gml.ac.cn, haishan.wu@zand.ae}

\begin{document}

\begin{frontmatter}
\maketitle

\begin{abstract}
Mangroves play a crucial role in maintaining coastal ecosystem health and protecting biodiversity. Therefore, continuous mapping of mangroves is essential for understanding their dynamics. Earth observation imagery typically provides a cost-effective way to monitor mangrove dynamics. However, there is a lack of regional studies on mangrove areas in the UAE. This study utilizes the UNet++ deep learning model combined with Sentinel-2 multispectral data and manually annotated labels to monitor the spatiotemporal dynamics of densely distributed mangroves (coverage greater than 70\%) in the UAE from 2017 to 2024, achieving an mIoU of 87.8\% on the validation set. Results show that the total mangrove area in the UAE in 2024 was approximately 9,142.21 hectares, an increase of 2,061.33 hectares compared to 2017, with carbon sequestration increasing by approximately 194,383.42 tons, equivalent to fixing about 713,367.36 tons of carbon dioxide. Abu Dhabi has the largest mangrove area and plays a dominant role in the UAE's mangrove growth, increasing by 1,855.6 hectares between 2017-2024, while other emirates have also contributed to mangrove expansion through stable and sustainable growth in mangrove areas. This comprehensive growth pattern reflects the collective efforts of all emirates in mangrove restoration.

% Please include a maximum of seven keywords
\keywords{Mangroves, Sentinel-2, Deep Learning, Dynamic Monitoring, United Arab Emirates}
\end{abstract}

\end{frontmatter}

\section{Introduction}
%%==================================%%
%%          INTRODUCTION            %%
%%==================================%%
%\section{Introduction}\label{sec1} 

% \textbf{The highest species richness across Mangrove ecosystems has been observed under moderate salinity \citep{Ball1998richness,islam2016species}. High species richness, in turn, is generally related to high productivity \citep{mittelbach2001observed,bai2021mangrove}, even though this relationship remains controversial and may depend on the spatial scale of interest \citep{whittaker2003observed,mensah2018diversity}. \\} 

\indent Mangrove ecosystems play a vital role in coastal environmental protection and biodiversity maintenance. However, mangroves around the world are facing serious degradation and loss. It is estimated that the global mangrove area has decreased by 25\%-50\% since 1980\citep{Spalding2022}. As one of the important countries with mangrove distribution on the Arabian Peninsula, the changes in the mangrove ecosystem in the UAE have attracted much attention. In recent years, with the development of remote sensing and geographic information system technology, research on the changes in the area of mangroves in the UAE has made some progress.

\indent Studies have shown that the area of mangroves in the UAE has fluctuated over the past few decades. On the one hand, urbanization and coastal development have led to the destruction of some mangroves\citep{Elmahdy2020,AlAbdulrazzak2019}; on the other hand, the protection and restoration measures taken by the government have increased the area of mangroves in some areas\citep{Elmahdy2020,Almahasheer2018}. Al Habshi et al .\citep{AlHabshi2011} used remote sensing data to analyze the distribution of mangroves in the UAE and found that mangroves are mainly distributed in coastal areas of emirates such as Abu Dhabi, Dubai and Sharjah. Moore et al. \citep{Moore2014} studied the distribution, pore water chemistry and community characteristics of mangroves in the UAE and found that the area of mangroves is about $41~\text{km}^2$ . Elmahdy et al . \citep{Elmahdy2020} used machine learning algorithms to monitor the changes in mangroves in the northern emirates from 1990 to 2019 and found that there was an overall growth trend. Alsumaiti et al. \citep{Alsumaiti2020} used Landsat satellite images to analyze the changes in mangroves on the eastern coast of Abu Dhabi from 1990 to 2014 and found that the area of mangroves in the region increased by about 35\%. They believe that this is mainly due to the protection and restoration plans of the local government.

\indent However, the latest trends and driving mechanisms of mangrove area changes in the UAE are still poorly understood. In particular, with the intensification of climate change and the continued advancement of coastal development in recent years, mangrove ecosystems may face new threats and opportunities, but there is a lack of systematic monitoring data. In addition, existing studies are mostly focused on individual emirates such as Abu Dhabi, and there are few comprehensive analyses of mangrove changes across the country.

\indent In recent years, deep learning technology has made significant breakthroughs in the field of remote sensing image classification and segmentation. With its powerful feature extraction and representation capabilities, convolutional neural networks (CNNs) have demonstrated superior performance in various remote sensing applications \citep{Zhu2017}  . In semantic segmentation tasks, U-Net architecture\citep{Ronneberger2015} , SegNet \citep{Badrinarayanan2017} , and DeepLab \citep{Chen2017} are widely used in high-resolution remote sensing image analysis. These models can effectively integrate multi-scale features and improve segmentation accuracy through encoder-decoder structures and jump connections. In the field of mangrove monitoring with remote sensing images, deep learning methods have also been widely applied. Li et al. \citep{Li2021} used an improved U-Net model to classify mangroves in Hainan Island, China, and achieved results that were superior to traditional machine learning methods. Pham et al . \citep{Pham2020}introduced the attention mechanism into U-Net to improve the accuracy of mangrove classification in Vietnam. 

\indent The UNet++ model\citep{Zhou2018}, as an improved deep learning semantic segmentation architecture, has achieved good performance in medical image segmentation and has gradually been applied to remote sensing image analysis in recent years \citep{Wang2022}. However, the application of UNet++ in mangrove remote sensing monitoring is still very limited. Therefore, the potential of applying UNet++ to mangrove extraction needs further evaluation, especially in the monitoring of mangroves in arid regions like the UAE, where no relevant reports have been published yet.

\indent This study aims to use the UNet++ model combined with Sentinel-2 multispectral data to monitor the spatiotemporal changes in mangrove forest of the UAE from 2017 to 2024. Specifically, the objectives of this study are: (1)Extract the spatial distribution of dense mangrove forests (coverage greater than 70\%) in the UAE and analyze their spatiotemporal trends; (2)to explore the main driving factors of mangrove changes. The major contribution of this study are: (1) Applying a deep learnig based workflow to large-scale monitoring of mangroves in the UAE based on the UNet++ model ; (2) Creating a 10-meter resolution mangrove dataset for the UAE in October 2020 using a semi-automated expert interpretation method based on the Global Mangrove Watch (GMW) v4.0 dataset combined with high-resolution satellite imagery from Google Earth; It is served as the training labels for the deep learning model. (3) revealing of the latest dynamics of mangrove area changes in the UAE. This study will provide a scientific basis for the formulation of effective mangrove protection and management strategies, which is of great significance for maintaining the health of the UAE coastal ecosystem.

\section{ Study Area and Materials}

\subsection{Overview of the Study Area}

\begin{figure}[t!]
\centering
\includegraphics[width=1.0\linewidth]{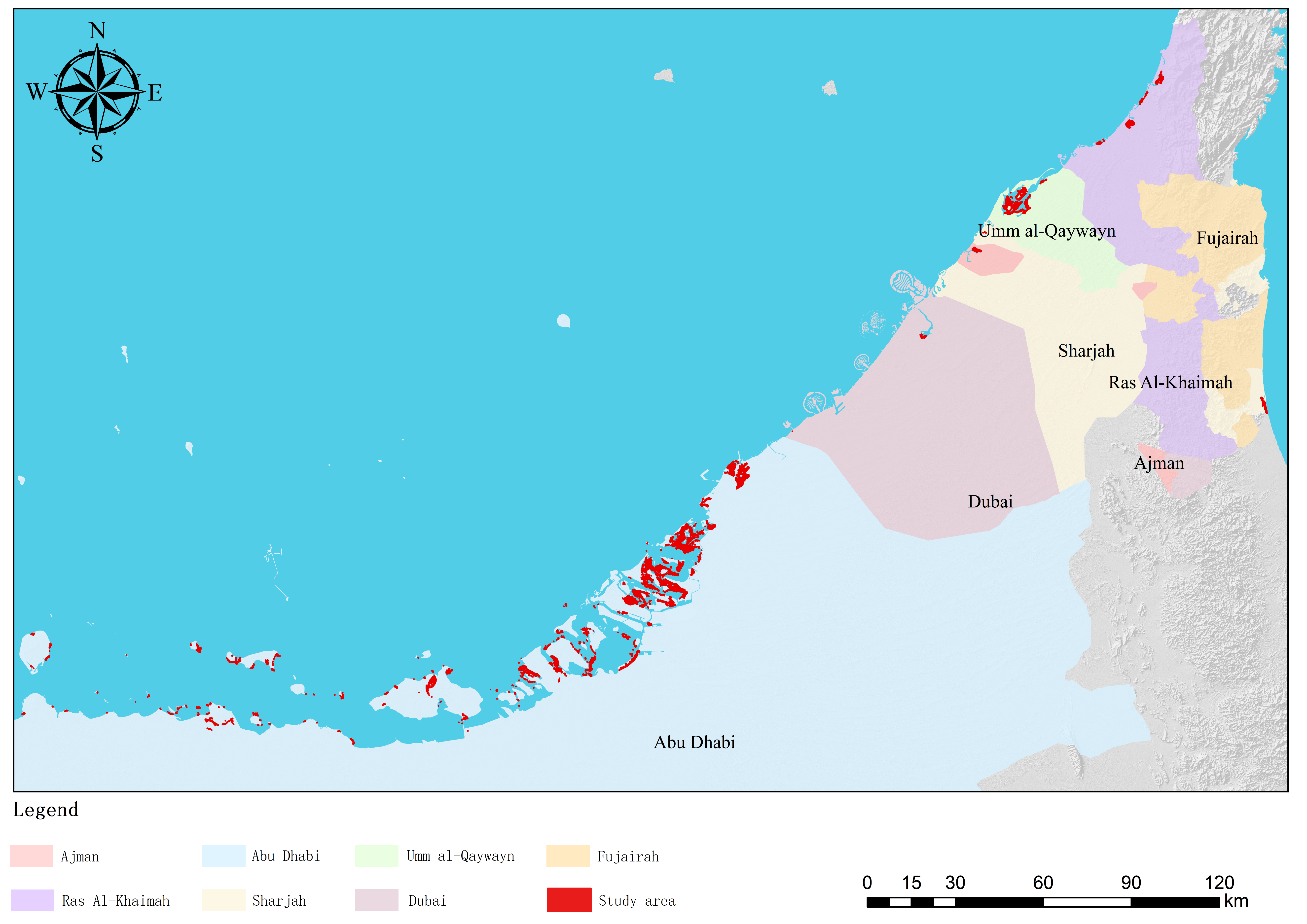}
\caption{UAE Mangrove Study Area}
\label{fig1}
\end{figure}

\indent The study area covers the entire United Arab Emirates (UAE), as shown in Fig~\ref{fig1}. The UAE is located in the eastern part of the Arabian Peninsula, bordered by the Persian Gulf to the north, the Gulf of Oman to the east, and sharing borders with Saudi Arabia and Oman to the south and west, respectively, with a coastline of approximately 1,318 kilometers\citep{Abdouli2019}. The study area can be roughly divided into two main parts: the northern emirates (including Dubai, Sharjah, Ajman, Umm Al-Quwain, and Ras Al Khaimah, with a total area of about $2{,}622~\text{km}^2$\citep{Elmahdy2022}) and the Emirate of Abu Dhabi. The UAE has a tropical desert climate, characterized by hot and dry weather throughout the year, with an average annual temperature of 28°C and annual precipitation typically ranging from 50 to 100 millimeters\citep{Boer1997}.

\indent The mangroves in the United Arab Emirates are primarily distributed in the intertidal zones along the coasts of the Persian Gulf and the Gulf of Oman, forming unique coastal wetland ecosystems. The main distribution areas include the eastern coastal region of Abu Dhabi, the Ras Al Khor Wildlife Sanctuary in Dubai, the Khor Kalba Nature Reserve in Sharjah, the Khor al Beidah lagoon area in Umm Al Quwain, and the coastal areas of Ras Al Khaimah. According to Alsumaiti's (2014) research, mangrove habitats in the UAE's coastal areas can be classified into three types: fringe, basin, and overwash \citep{Alsumaiti2014}. Avicennia marina is the primary native mangrove species in the UAE's coastal areas, capable of tolerating high soil salinity, high environmental temperatures, and limited freshwater influx \citep{Elmahdy2020}. The height of mangroves in the UAE ranges from a few centimeters to 3-8 meters \citep{Alsumaiti2014,Moore2015}. These mangroves play a crucial role in coastal ecosystems. This study will focus on analyzing the changes in mangrove area across different emirates of the UAE, exploring various factors influencing these changes, and providing scientific basis for the conservation and management of mangroves in the UAE.

\subsection{Data Sources and Preprocessing}

\subsubsection{Remote Sensing Image Data}

\indent This study utilizes Sentinel-2 multispectral satellite imagery as remote sensing data, spanning from 2017 to 2024. The Sentinel-2 satellite was launched on June 23, 2015, equipped with a Multi-Spectral Instrument (MSI) that captures 13 spectral bands with spatial resolutions ranging from 10 to 60 meters. The visible and near-infrared bands have a resolution of 10 meters. Sentinel-2 data was chosen primarily for its high temporal resolution (5-day revisit cycle), high spatial resolution, and rich spectral information, making it particularly suitable for large-scale, long-term time series studies of mangrove forests with fine detail. All Sentinel-2 image data were acquired and pre-processed through the Google Earth Engine (GEE) platform. GEE is a cloud-based geospatial analysis platform that provides access to massive earth observation satellite data and powerful data processing capabilities.

\indent The data pre-processing workflow is as follows: First, Sentinel-2 Level-2a data with cloud coverage less than 10\% between 2017 and 2024 were selected. Subsequently, cloud removal was performed using the quality assessment band. To minimize the impact of tides on mangrove extraction, we used the Normalized Difference Water Index (NDWI) to extract water bodies in the study area, calculated the water area for each month of every year, and then selected the two months with the smallest water area as the time window for image filtering. Images within the selected months were mosaicked and clipped to the study area extent. This method ensures that the acquired images maximally reflect the actual distribution of mangroves while minimizing the effects of tides and cloud cover. Through the GEE platform, we can quickly access and process large volumes of Sentinel-2 images, greatly improving the efficiency of data acquisition and pre-processing. This cloud computing-based approach enables us to efficiently process large amounts of spatiotemporal data, providing a reliable data foundation for subsequent mangrove change monitoring.

\subsubsection{Label Data Acquisition and Optimization}

\indent This study uses the 2020 mangrove distribution data released by the Global Mangrove Watch (GMW) v4.0 as the base labeled dataset\citep{GlobalMangroveWatch2022}. This dataset was processed using Sentinel-2 satellite imagery with a 10-meter resolution to remap the mangrove distribution, including many areas previously unmapped. Compared to the former 25-meter pixel resolution, the new dataset’s 10-meter resolution allows for more detailed mapping of mangrove extent, capturing finer features such as fringing and riverine mangroves. As the latest and most comprehensive global mangrove distribution dataset available, the GMW dataset provides a reliable baseline for large-scale mangrove mapping. However, considering that some local inaccuracies may exist in the GMW data, this study used high-resolution satellite imagery from 2020 on the Google Earth platform to systematically refine and visually interpret mangrove distributions in the study area. The interpretation process focused on the typical characteristics of mangroves (such as canopy texture and tone) and the precise delineation of boundary areas.

\indent After aligning the optimized label data with Sentinel-2 imagery, the data was cropped into 256×256 pixel sizes to generate the final training dataset. This dataset comprises a total of 1,568 training samples, including 346 images containing mangroves (22.07\%) and 1,222 images without mangroves (77.93\%). This approach, which combines GMW data with manual optimization based on high-resolution imagery, ensures both the systematic nature and reliability of the label data while improving the accuracy in local areas. It provides high-quality annotated data for training deep learning models.

\section{Methodology}

\subsection{UNet++ Model Architecture}
\begin{figure}[htbp]
\centering
\includegraphics[width=1.0\columnwidth]{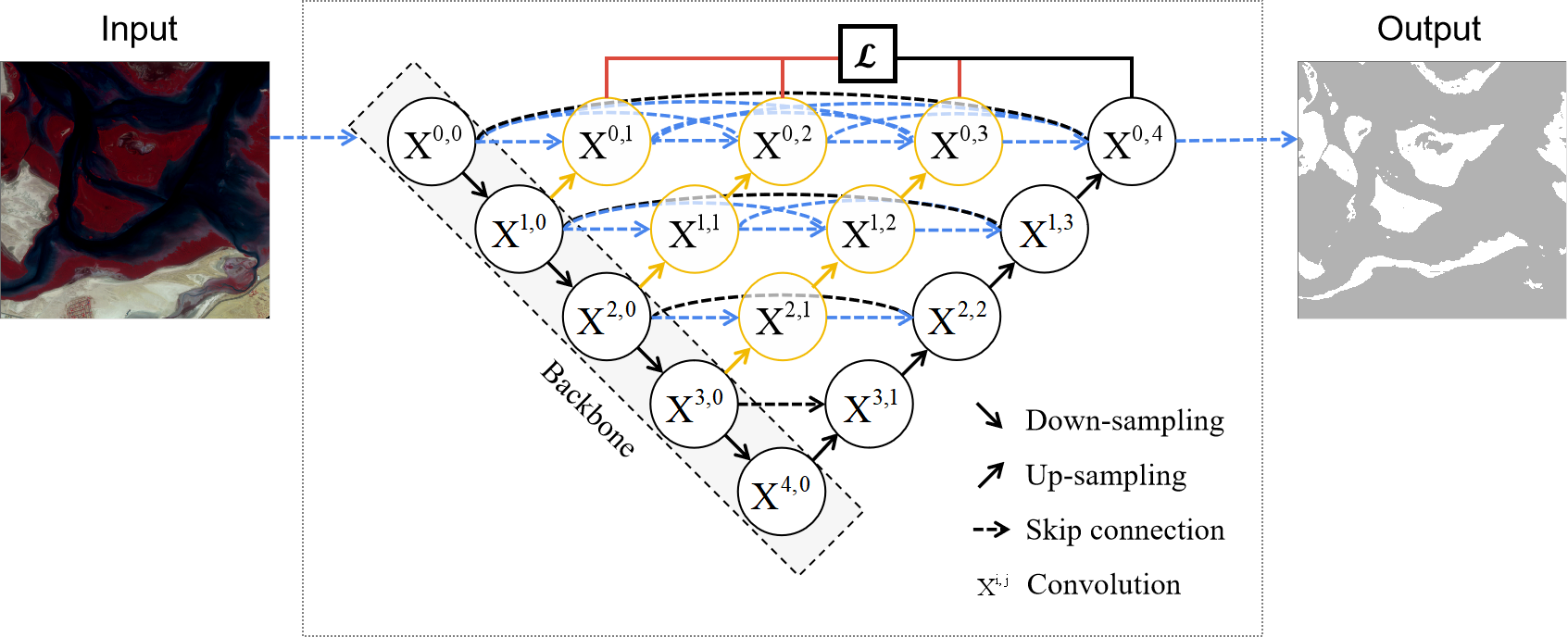}
\caption{UNet++ model architecture}
\label{fig2}
\end{figure}

\indent The UNet++ model is an improved version of the classic UNet, optimizing feature extraction and fusion processes through redesigned skip connections and a dense nested structure \citep{Zhou2018}. The network structure is shown in Fig~\ref{fig2}. The core innovation of this model lies in the introduction of dense skip pathways, which reduce the semantic gap between encoder and decoder through deep fusion of multi-scale features, thereby improving segmentation accuracy. In this study, UNet++ was chosen as the basic framework mainly based on the following considerations: First, mangroves often present complex boundary features and spatial heterogeneity in remote sensing images, requiring the model to have strong feature extraction and boundary recognition capabilities. Second, the dense skip connection structure helps to preserve and fuse spatial information at different scales, which is crucial for accurately identifying mangrove patches of varying sizes.

\subsection{Model Implementation}

\indent The input of the UNet++ model consists of 11 multispectral bands from Sentinel-2, with the output being binary classification results (mangrove/non-mangrove). For dataset partitioning, 10\% (35 images) were randomly selected from the 346 images containing mangroves to form the validation set. The remaining 311 images, along with 1222 images without mangroves, comprise the training set, totaling 1533 images.

\indent This study employs timm-resnest101e as the encoder backbone for UNet++, incorporating ImageNet pre-trained weights to enhance the model's feature extraction capabilities. The loss function combines SoftCrossEntropyLoss and DiceLoss to address class imbalance and improve boundary segmentation accuracy. SoftCrossEntropyLoss enhances the model's generalization ability through label smoothing, while DiceLoss focuses on optimizing segmentation accuracy. The AdamW optimizer is used with an initial learning rate of 0.0001 and a weight decay of 0.001. To prevent model overfitting and improve training efficiency, a cosine annealing strategy is adopted to dynamically adjust the learning rate, with the initial restart period \( T_0 \) set to 2, the multiplier \( T_{\text{mult}} \) set to 2, and the minimum learning rate set to 0.00001.

\subsection{Training Strategy Optimization}

\begin{figure}[htbp]
\centering
\includegraphics[width=1.0\columnwidth]{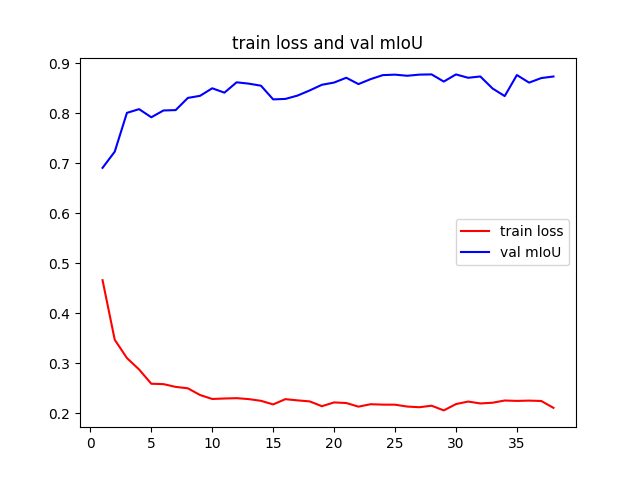}
\caption{Validation set mIoU and training set loss curves during UNet++ model training}
\label{fig3}
\end{figure}

\indent The model training employs a mini-batch iteration approach, with the batch size set to 16 to balance computational efficiency and model performance. The training process is set for 100 epochs, but an early stopping mechanism is introduced, halting training when the validation set's mIoU shows no improvement for 10 consecutive epochs. To enhance model robustness, data augmentation techniques are applied during training, including random flipping and rotation operations. Additionally, considering the uneven distribution of mangroves in the study area, a spatially stratified sampling strategy is adopted to construct training batches, ensuring the model can adequately learn mangrove features from different regions. Throughout the training process, the model's performance is monitored using the mIoU metric on the validation set, and the best model weights are saved for subsequent predictions.Fig~\ref{fig3} illustrates the changes in validation set mIoU and training set loss curves during the UNet++ model training process. Throughout the entire training process, the model's loss value and validation set mIoU demonstrate a gradual convergence trend. The training loss decreases rapidly within the first 15 epochs and then stabilizes, indicating good model convergence. As training progresses, the mIoU gradually improves, ultimately reaching its best performance at the 28th epoch with an mIoU of 87.8\%.

\indent This optimized network architecture design and training strategy enable the model to effectively address key challenges in mangrove extraction, such as complex spatial distribution patterns and blurred boundaries. The final achievement of 87.8\% mIoU on the validation set demonstrates the effectiveness of this method in mangrove remote sensing mapping.

\section{Results}
\subsection{Accuracy Evaluation}
\indent This study primarily focuses on the extraction accuracy of mangroves as the target category, therefore employing user accuracy, producer accuracy (recall), and F1 score as evaluation metrics. The reason for choosing these two evaluation metrics instead of overall accuracy is that the research emphasis is on the extraction effectiveness of mangroves as a specific land cover type, and non-mangrove pixels account for as high as 77.93\% of the dataset. Using overall accuracy might obscure the actual performance of mangrove extraction.

\begin{figure}[t!]
\centering
\includegraphics[width=1.0\columnwidth]{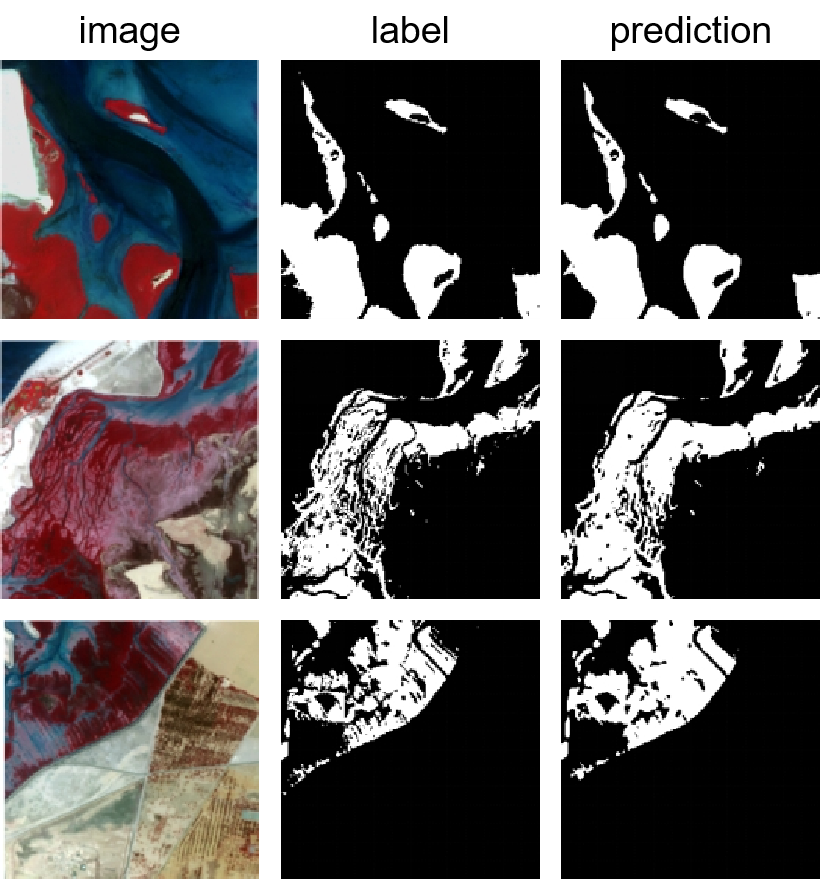}
\caption{Mangrove Extraction Results of the UNet++ Model on the Test dataset ( optical satellite image on the left, the real mangrove area map in the middle, and the predicted mangrove area map on the right )}
\label{fig4}
\end{figure}

\indent We selected the 2024 mangrove distribution data as the test dataset. This data was derived from the 2020 mangrove distribution data, updated through manual visual interpretation using high-resolution Google Earth imagery to ensure the accuracy and timeliness of the test data. According to the experimental results, the model achieved excellent performance in mangrove extraction, with a producer accuracy (recall) of 0.9196, meaning the model correctly identified 91.96\% of the actual mangrove pixels. The user accuracy was 0.8917, indicating that 89.17\% of the pixels predicted as mangroves by the model were indeed mangroves. Additionally, the F1 score reached 0.9055, reflecting a good balance between producer accuracy and user accuracy. From the prediction result shown in Fig~\ref{fig4}, it can be visually observed that the model has a good ability to capture the boundary and morphological characteristics of mangroves. These metrics collectively demonstrate that the UNet++ model can effectively identify and extract mangrove areas, showing strong practical value.

\subsection{Analysis of Mangrove Changes in the UAE (2017-2024)}
\begin{figure}[htbp]
\centering
\includegraphics[width=1.0\columnwidth]{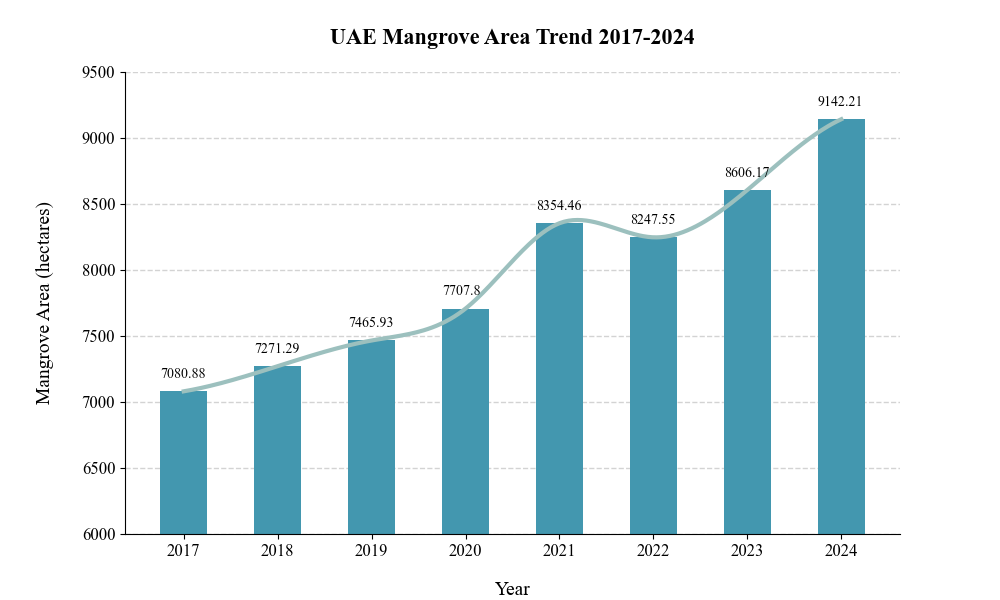}
\caption{Changes in mangrove area in the UAE from 2017 to 2024}
\label{fig5}
\end{figure}

\indent This study analyzed remote sensing imagery to obtain data on mangrove area in the United Arab Emirates (UAE) from 2017 to 2024. The results show a significant increasing trend in the total mangrove area in the UAE during this period (Fig~\ref{fig5}), growing from 7,080.88 hectares in 2017 to 9,142.21 hectares in 2024 (Tab~\ref{tab1}), representing a total increase of 29.11\%. According to research, mangroves represent the blue carbon system with the highest carbon storage potential. With the UAE's mangroves having an average carbon density of approximately 94.3 tons of carbon per hectare \citep{Carpenter2023}, it can be estimated that the carbon sequestration by UAE mangroves increased by 194,383.42 tons of carbon from 2017 to 2024, equivalent to fixing about 713,367.36 tons of carbon dioxide. This growth reflects the UAE's significant achievements in mangrove protection and restoration efforts.

\indent Calculations indicate that the average annual growth rate of mangrove area in the UAE between 2017 and 2024 was approximately 3.71\%. This sustained growth trend reflects the effective measures taken by the UAE in mangrove conservation and restoration.Despite the overall growth trend, the rate of increase was not uniform. The data shows that there were several periods where the growth rate was significantly higher than the average: between 2020-2021, the area increased by 646.66 hectares, with an annual growth rate of 8.39\%, the highest within the study period; between 2023-2024, the area increased by 539.04 hectares, with an annual growth rate of 6.23\%. Growth was slow from 2017-2020, with rates below the average. In contrast, there was a slight decrease in 2021-2022, with an area reduction of 106.91 hectares, representing a 1.28\% decline. The reduction was attributed to the construction of artificial waterways, which can be directly observed from satellite imagery in Um Yifeenah Island, Abu Dhabi, demonstrating how human activities directly led to the decrease in mangrove area. Other potential factors contributing to the area reduction include climate change or natural disturbances. Notably, although the UAE's mangrove area experienced a temporary decrease, it has maintained an overall stable growth trend, reflecting the UAE's continuous efforts in protecting mangrove ecosystems.

\begin{table}[htbp]
\caption{Statistics of mangrove area in the UAE}
\centering
\begin{tabular}{|c|c|c|c|}
\hline
\textbf{Years} & \textbf{Area (ha)} & \textbf{Annual change (ha)} & \textbf{Annual change rate (\%)} \\
\hline
2017 & 7080.88 & & \\
2018 & 7271.29 & 190.41 & 2.69 \\
2019 & 7465.93 & 194.64 & 2.68 \\
2020 & 7707.80 & 241.87 & 3.24 \\
2021 & 8354.46 & 646.66 & 8.39 \\
2022 & 8247.55 & -106.91 & -1.28 \\
2023 & 8606.17 & 358.62 & 4.35 \\
2024 & 9142.21 & 539.04 & 6.23 \\
\hline
\end{tabular}
\label{tab1}
\end{table}

\subsection{Analysis of Changes in Mangrove Area in Each Emirate}

\begin{figure}[htbp]
\centering
\includegraphics[width=1.0\columnwidth]{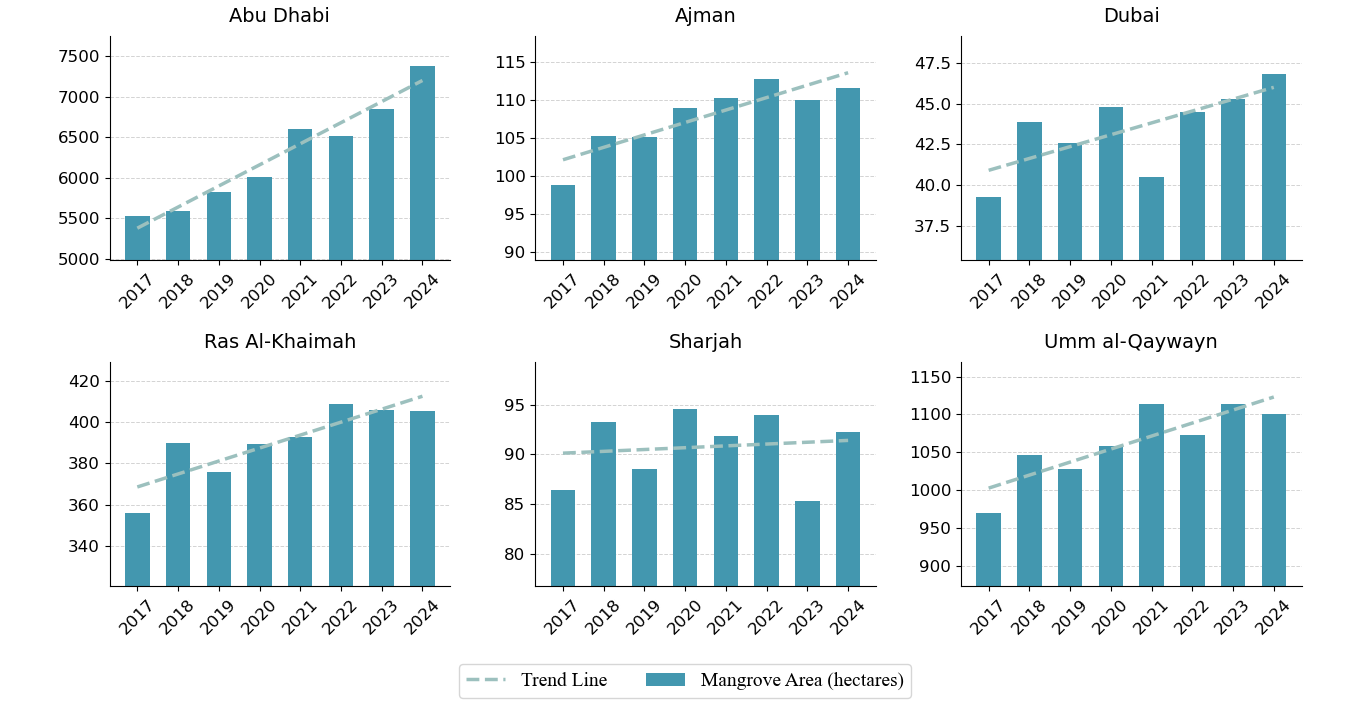}
\caption{Mangrove area and changing trends in each emirate of the UAE}
\label{fig6}
\end{figure}

\indent The research results show that only the Emirate of Fujairah has no mangrove distribution. This may be due to Fujairah lacking a suitable coastal environment for mangrove growth, or not having initiated any mangrove planting projects yet. The remaining six emirates generally see an increase in mangrove area, but the patterns of change and the extent of increase vary significantly among the emirates ( Fig~\ref{fig6}). The following provides a detailed analysis of the changes in mangrove area from 2017 to 2024 for these six emirates.

\indent Abu Dhabi, as the emirate with the largest mangrove area in the UAE, shows an overall growth trend, though not linear. It increases from 5,530.08 hectares in 2017 to 7,385.68 hectares in 2024, an increase of 1,855.6 hectares, representing a total growth of 33.5\%. However, year-to-year changes show fluctuations. Only the period from 2021 to 2022 shows a slight decrease, with the area dropping from 6,563.75 hectares to 6,475.43 hectares. Other years maintain a significant growth trend, reaching the highest value of the observation period in 2024.

\indent Ajman's mangrove area generally shows an upward trend, increasing from 98.83 hectares in 2017 to 111.52 hectares in 2024, an increase of 12.69 hectares, representing a total growth of 12.8\%. The most significant growth occurs in 2017-2018, increasing by 6.44 hectares. From 2019 to 2022, it shows continuous small growth, while other years see slight decreases or remain relatively stable.

\indent Dubai's mangrove area shows an overall fluctuating growth trend. It increases from 39.28 hectares in 2017 to 46.84 hectares in 2024, an increase of 7.56 hectares, representing a total growth of 19.2\%. The 2017-2018 period sees a larger increase of 4.62 hectares, and from 2021 to 2024, it shows continuous slow growth. The 2020-2021 period sees a larger decrease of 4.28 hectares. In other years, the mangrove area remains relatively stable, with annual changes around 2 hectares.

\indent Ras Al Khaimah's mangrove area also shows an overall fluctuating growth trend. It increases from 356.21 hectares in 2017 to 405.40 hectares in 2024, an increase of 49.19 hectares, representing a total growth of 13.8\%. The most significant growth occurs in 2017-2018, increasing by 33.58 hectares. From 2019 to 2022, it shows continuous growth, reaching a peak of 408.72 hectares in 2022. However, there is a relatively large decrease of 13.83 hectares in 2018-2019, and after 2022, the mangrove area remains relatively stable.

\indent Sharjah's mangrove area shows a trend of initial increase, followed by a decrease, and then an increase again from 2017 to 2024, with an overall slow growth. It increases from 86.39 hectares in 2017 to 92.3 hectares in 2024, an increase of 5.91 hectares, representing a total growth of about 6.91\%. In 2018, 2020, and 2024, the annual growth of mangrove area is around 6 hectares, showing slow growth. In 2023, there is a larger decrease of 8.64 hectares, while other years show little change.

\indent The mangrove area in Umm Al Quwain shows an overall growth trend from 2017 to 2024, despite annual fluctuations, increasing by 130.38 hectares from 2017 to 2024. The period from 2017 to 2024 can be roughly divided into two stages: steady growth (2017-2021) and fluctuating decline (2021-2024). From 2017 to 2021, there is a cumulative increase of 143.58 hectares, showing a significant growth, although there is a slight decrease between 2018 and 2019. After reaching a peak in 2021, it enters a stage of small fluctuations, with little overall change. From 2021 to 2024, the mangrove area changes from 1,113.67 hectares to 1,100.47 hectares.

\subsection{Analysis of Spatial Distribution Changes of Mangroves}

\begin{figure}[t!]
\centering
\includegraphics[width=1.0\columnwidth]{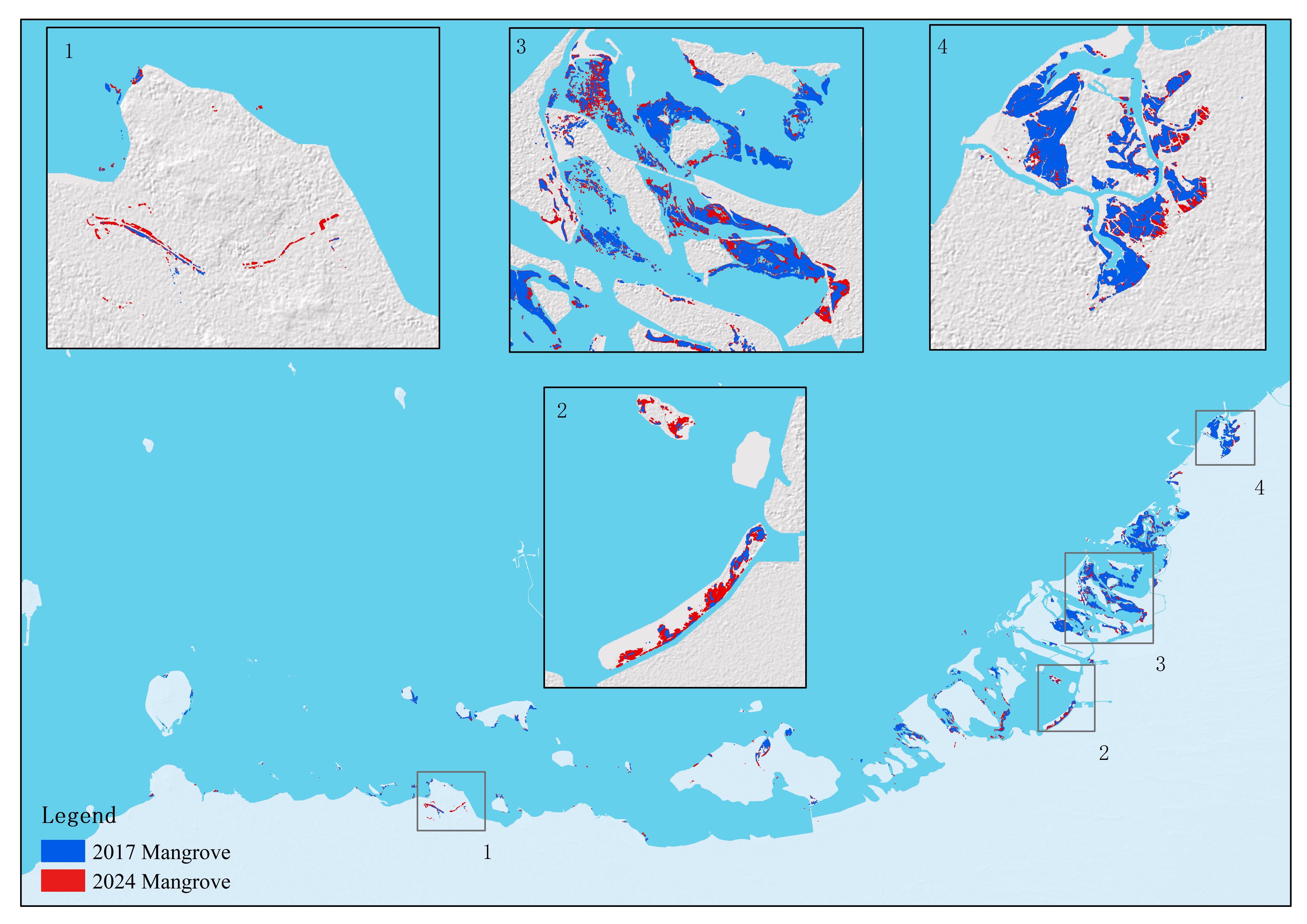}
\caption{Spatial changes in mangrove forests in the Emirate of Abu Dhabi}
\label{fig7}
\end{figure}

\indent Fig~\ref{fig7} shows the relative spatial distribution changes in Abu Dhabi from 2017 to 2024. The blue areas represent the mangrove coverage in 2017, while the red areas represent the mangrove coverage in 2024. Numbers 1, 2, 3, and 4 indicate enlarged areas where changes are particularly noticeable.

\indent Area 1 is located at Qareen Al Aish, characterized by a complex coastline with numerous inland waterways, lagoons, and small islands. This intricate water system near Qareen Al Aish provides highly favorable conditions for mangrove growth and expansion. Between 2017 and 2024, the mangrove coverage in this area significantly increased, not only filling gaps between existing mangroves but also expanding into new suitable areas. The growth areas are often in regular rectangular shapes, indicating clear signs of artificial planting.

\indent Area 2 is near Al Hudayriat Island. In 2017, mangrove distribution in this area was sparse with limited coverage. By 2024, the mangrove area has significantly increased, forming a denser and more continuous vegetation belt. The curved coastline on the southeast side of the island shows the most notable mangrove expansion.

\indent Area 3 is near Al-Jubail Island, with notable growth on Al-Jubail Island, Fahid Island, Zeraa Island, and Yas Island. The mangrove coverage around Al-Jubail Island shows the most obvious growth from 2017 to 2024. In 2017, there were only scattered mangroves around the island, mainly distributed along the southern and eastern coasts. However, by 2024, the mangrove coverage has expanded significantly, with new areas mainly concentrated in the north, west, and southwest of the island, while the original mangrove areas in the east have also further expanded. This significant growth pattern is reflected not only in the increase in area but also in the extensiveness and density of distribution. Such remarkable changes may be the result of mangrove protection and planting programs implemented in the area, effectively promoting the restoration and expansion of mangrove ecosystems.On Fahid Island, there was a small distribution of mangroves in 2017, mainly in the south and west of the island. By 2024, the mangrove area has significantly increased, especially in the west and north of the island. The distribution of mangroves around the island is more uniform and dense.Around Zeraa Island, mangroves were sparsely distributed in 2017, mainly concentrated in the north and east of the island. By 2024, mangroves have almost covered the entire periphery of the island. Particularly in the south and west of the island, a large number of new mangroves have been added.On the northeast part of Yas Island, there was some mangrove distribution in 2017. By 2024, the mangrove area has significantly increased, mainly concentrated in the northeast and east of the island. The newly added mangrove areas have clearly expanded, forming a more continuous strip-like distribution.

\begin{figure}[htbp]
\centering
\includegraphics[width=1.0\columnwidth]{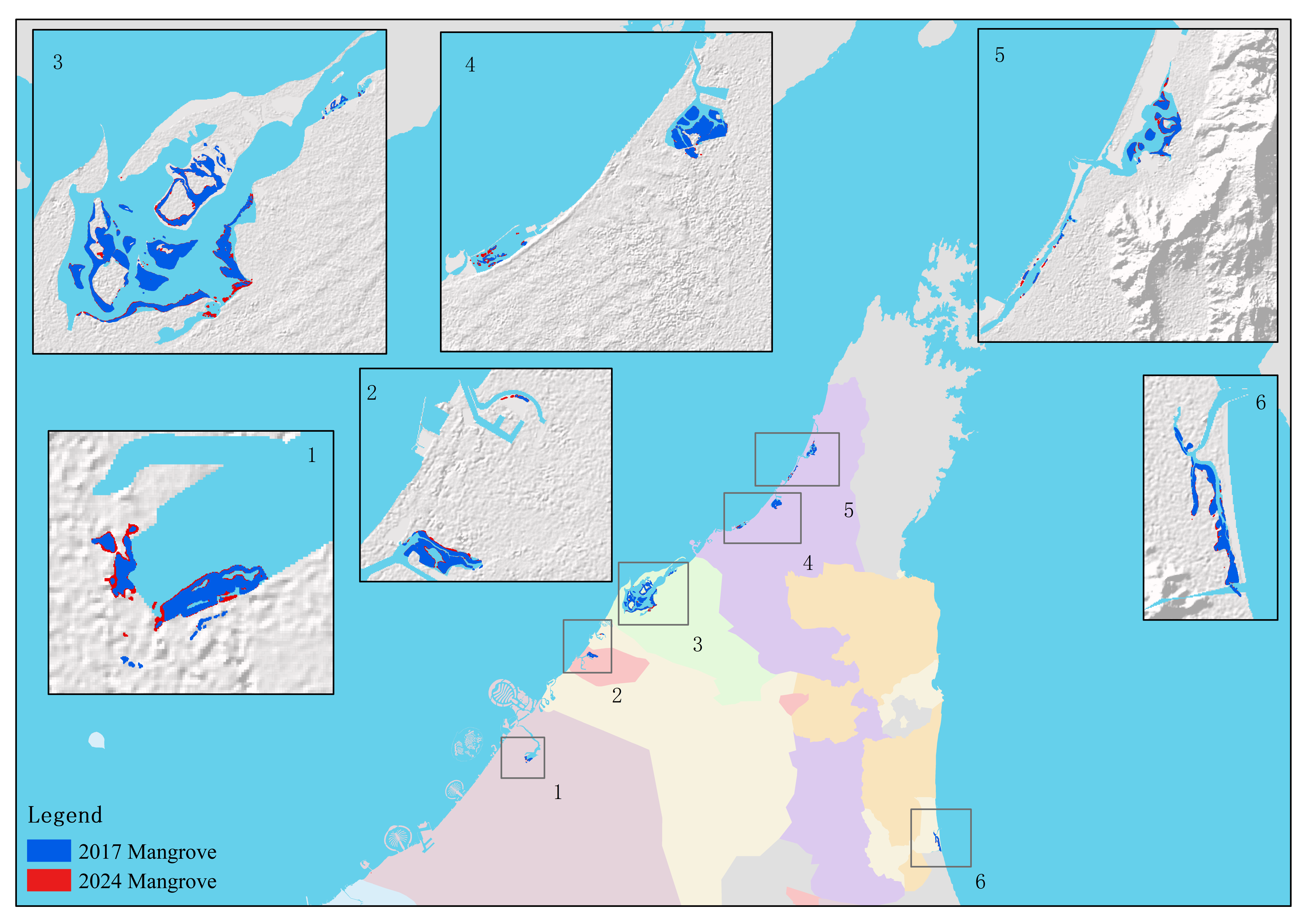}
\caption{Spatial changes in mangrove forests in the northern emirates of the UAE}
\label{fig8}
\end{figure}

\indent Area 4 in Fig~\ref{fig8} is centered on Jazirat Taweelah Island. The mangrove coverage in Jazirat Taweelah and its surrounding coastal areas has also shown a significant growth trend in recent years. The newly added areas are mainly concentrated along the eastern and southern coasts of Jazirat Taweelah Island. This significant growth pattern is reflected not only in the increase in area but also in the improvement of distribution continuity and density. The mangrove distribution is gradually spreading from the coast towards inland areas.

\indent The mangrove ecosystems in the northern emirates of the UAE have also demonstrated positive conservation outcomes and steady development. According to Fig~\ref{fig8}, the mangrove distribution in Dubai's Ras Al Khor Wildlife Sanctuary and its surrounding areas shows an expansion trend outward (Area 1). In Sharjah and Ajman (Area 2), the distribution range of mangroves has also increased between 2017 and 2024. The mangrove distribution in Umm Al Quwain (Area 3) shows steady growth with incremental increases. The mangrove ecosystems in Ras Al Khaimah (Area 4 and Area 5) and Sharjah (Area 6) have also maintained a healthy state, keeping a stable growth pattern. Overall, the northern emirates have achieved positive results in the protection and development of mangrove ecosystems, laying a solid foundation for the sustainable development of coastal ecosystems.

\subsection{ Satellite Image Analysis of Typical Expansion Areas of Mangroves in Abu Dhabi}

\begin{figure}[htbp]
\centering
\includegraphics[width=1.0\columnwidth]{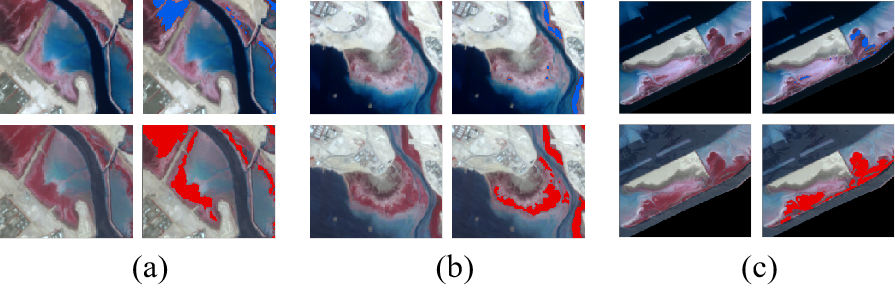}
\caption{Comparison of Sentinel-2 satellite images of typical mangrove expansion areas in Abu Dhabi in 2017 and 2024}
\label{fig9}
\end{figure}

\indent To more intuitively demonstrate the significant expansion of mangroves in Abu Dhabi between 2017 and 2024, this study analyzed false-color composite Sentinel-2 satellite images and mangrove extraction results for three typical growth areas (Fig~\ref{fig9}). Area a is located along the western coastal zone of Samaliyah Island. Satellite imagery from 2017 shows that this area was primarily exposed tidal flats. The 2024 satellite imagery reveals that mangroves in this area have gradually expanded into suitable zones, with a notable increase in coverage area. Area b, near the American Community School of Abu Dhabi, shows scattered mangrove distribution in 2017, but by 2024, coverage has significantly increased, forming a denser and more continuous vegetation belt. Area c, located near Zayed Military University, displays bright red mangroves and dark red salt marsh vegetation in the satellite images, with both images and interpretation results indicating a significant mangrove expansion trend. These changes reflect the improvement of Abu Dhabi's ecological environment and the effectiveness of local government protection measures.

\subsection{Analysis of Driving Factors for Mangrove Area Growth}

\indent The series of mangrove protection and restoration plans implemented by the UAE government is a key factor driving the growth of mangrove areas. The Environment Agency - Abu Dhabi (EAD) has set an ambitious goal of planting 100 million mangroves by 2030 \citep{NewScientist2023}. This commitment reflects the UAE's determination to address climate change and enhance the resilience of coastal ecosystems. According to reports, the project is using specially designed drones to disperse mangrove seeds, with these drones capable of scattering 2,000 seeds per minute. This efficient planting method is expected to significantly accelerate the speed of mangrove restoration. Furthermore, the project utilizes artificial intelligence technology to identify areas most suitable for mangrove growth, thereby improving the success rate of planting. The "Blue Carbon" project, launched by the UAE government in 2013, marks an important milestone in the country's recognition and utilization of the carbon sequestration capacity of marine ecosystems\citep{UAEBGC2022}. This project not only demonstrates the government's profound understanding of the ecological value of mangroves but has also directly promoted mangrove area growth in several key regions. The implementation of the "Blue Carbon" project has brought particularly significant results in the Emirate of Abu Dhabi. A decree issued in 2020 prohibits damage to mangroves, with violators facing fines of up to 10,000 dirhams, providing strong legal protection \citep{TheNational2020}. These policy measures have provided robust institutional support for the growth of mangrove areas, which may explain the sustained growth trend observed in this study, especially in the Abu Dhabi region.

\indent Mangroves demonstrate significant adaptability to local climate changes, which may be an important reason for their area expansion. The developed root systems of mangroves can survive in high-salinity environments, and their special leaf structures can reduce water evaporation \citep{ReefResilience2021}. They have strong resistance to strong winds and storm surges, enabling them to adapt to the frequent sandstorms in Abu Dhabi. The latest research by Alothman et al. (2023) found that mangroves in the Arabian Gulf, especially Avicennia marina, have developed unique physiological adaptations that allow them to thrive in environments of extreme heat and high salinity, enabling them to survive and expand in the UAE's harsh climate conditions \citep{Alothman2023}. This adaptability is particularly important against the backdrop of rising temperatures and sea levels in the region.

\section{Conclusion}
\indent The application of the UNet++ network model in this study provides an effective means for automated mangrove identification and monitoring. Through nested skip connections and deep feature fusion, UNet++ successfully addressed the complex boundary issues of mangroves and significantly improved segmentation accuracy. The validation set results of this study show that the UNet++ model achieved an mIoU of 87.8\% in mangrove extraction.

\indent Although this study only extracted dense mangrove areas with coverage greater than 70\%, the results still indicate significant mangrove expansion, demonstrating that the UAE's mangrove density has continued to increase in recent years with vigorous growth. Between 2017-2024, dense mangrove areas increased by approximately 2,061.33 hectares, equivalent to sequestering 194,383.42 tons of carbon, fixing about 713,367.36 tons of carbon dioxide. Abu Dhabi, being the largest emirate in the UAE and the main region for mangrove distribution, has made the primary contribution to mangrove growth. The large-scale mangrove planting projects implemented in Abu Dhabi in recent years have directly promoted the expansion of UAE's mangroves. Areas such as Qareen Al Aish, Al Hudayriat Island, and Al-Jubail Island in Abu Dhabi have shown evident patterns of mangrove expansion, with notably increased mangrove density, especially within protected areas and vegetation restoration zones. Other emirates have also shown stable growth trends and have made efforts in mangrove protection and restoration. The UAE has achieved remarkable results in mangrove restoration projects, setting an example for global mangrove ecosystem restoration.

\indent By combining Sentinel-2 multispectral data with the UNet++ model, this study successfully achieved large-scale, long-term dynamic monitoring of mangroves, providing the latest spatiotemporal data support for UAE's mangrove protection. The research findings not only provide scientific basis for developing more targeted mangrove protection and restoration policies, but also offer effective reference methods for remote sensing monitoring of mangroves in other arid climate regions, contributing to the promotion of global mangrove ecosystem protection and management.

\section*{Conflict of interest}
The authors declare that they have no competing financial interests.

% \section*{Supporting Information}

% Supporting information is information that is not essential to the article, but provides greater depth and background. It is hosted online and appears without editing or typesetting. It may include tables, figures, videos, datasets, etc. More information can be found in the journal's author guidelines or at \url{http://www.wileyauthors.com/suppinfoFAQs}. Note: if data, scripts, or other artefacts used to generate the analyses presented in the paper are available via a publicly available data repository, authors should include a reference to the location of the material within their paper.

\printendnotes

% Submissions are not required to reflect the precise reference formatting of the journal (use of italics, bold etc.), however it is important that all key elements of each reference are included.
%\bibliography{sample}
%
%\begin{biography}
%
%\end{biography}

% \graphicalabstract{example-image-1x1}{Please check the journal's author guildines for whether a graphical abstract, key points, new findings, or other items are required for display in the Table of Contents.}

\end{document}